\documentclass[10pt,twocolumn,letterpaper]{article}

\usepackage{cvpr}              
\usepackage[table]{xcolor}

\usepackage{pifont}
\newcommand{\cmark}{\textcolor{ForestGreen}{\ding{51}}} 
\newcommand{\xmark}{\textcolor{red}{\ding{55}}}         
\usepackage{multirow}

\definecolor{cvprblue}{rgb}{0.21,0.49,0.74}
\usepackage[pagebackref,breaklinks,colorlinks,allcolors=cvprblue]{hyperref}
\usepackage{bbding}
\def\paperID{3216} 
\def\confName{CVPR}
\def\confYear{2026}

\title{FreeGen: Feed-Forward Reconstruction–Generation Co-Training for Free-Viewpoint Driving Scene Synthesis}

\author{
Shijie Chen \quad Peixi Peng \textsuperscript{\Envelope}\\
School of Electronic and Computer Engineering, Peking University\\
{\tt\small \{2501111941@stu., pxpeng\}@pku.edu.cn}
}

\begin{document}
\maketitle
\begin{abstract}
Closed-loop simulation and scalable pre-training for autonomous driving require synthesizing free-viewpoint driving scenes. However, existing datasets and generative pipelines rarely provide consistent off-trajectory observations, limiting large-scale evaluation and training. While recent generative models demonstrate strong visual realism, they struggle to jointly achieve interpolation consistency and extrapolation realism without per-scene optimization. To address this, we propose FreeGen, a feed-forward reconstruction-generation co-training framework for free-viewpoint driving scene synthesis. The reconstruction model provides stable geometric representations to ensure interpolation consistency, while the generation model performs geometry-aware enhancement to improve realism at unseen viewpoints. Through co-training, generative priors are distilled into the reconstruction model to improve off-trajectory rendering, and the refined geometry in turn offers stronger structural guidance for generation. Experiments demonstrate that FreeGen achieves state-of-the-art performance for free-viewpoint driving scene synthesis.
\end{abstract}    
\section{Introduction}
\label{sec:intro}
Robust autonomous driving systems require virtual environments for scalable pretraining and closed-loop policy evaluation~\cite{hu2022st, hu2023planning, jiang2023vad}. However, existing real~\cite{caesar2020nuscenes, sun2020scalability, geiger2012we} and synthetic driving data~\cite{gao2023magicdrive, gao2025magicdrive, wang2024drivedreamer} are predominantly collected along a single trajectory, providing limited coverage of the diverse trajectories induced by different actions. The scarcity of off-trajectory observations hinders high-fidelity and consistent rendering at free-viewpoints~\cite{yan2024street, chen2023periodic, yang2023emernerf, chen2024omnire}.
The key challenge lies in achieving high-quality free-viewpoint driving scene synthesis across different trajectories using only single-trajectory observations.

\begin{table}[!ht]
\caption{\textbf{Comparison of different methods.} We consider interpolation consistency (Interp. Consistency), extrapolation realism (Extrap. Realism), and whether supports feed-forward inference.}
\vspace{-1ex}
\centering
\setlength{\tabcolsep}{4pt}
\resizebox{\linewidth}{!}{
\begin{tabular}{l|ccc}
\toprule
Method & Interp. Consistency & Extrap. Realism & Feed-forward \\
\midrule
3DGS~\cite{kerbl20233d}            & \cmark & \xmark & \xmark \\
PVG~\cite{chen2023periodic}              & \cmark & \xmark & \xmark \\
UniSim~\cite{yang2023unisim}        & \cmark & \xmark & \xmark \\
EmerNeRF~\cite{yang2023emernerf}    & \cmark & \xmark & \xmark \\
StreetGaussian~\cite{yan2024street} & \cmark & \xmark & \xmark \\
OmniRe~\cite{chen2024omnire} & \cmark & \xmark & \xmark \\
\midrule
DriveDreamer4D~\cite{zhao2025drivedreamer4d}        & \cmark & \cmark & \xmark  \\
ReconDreamer~\cite{ni2025recondreamer}          & \cmark & \cmark & \xmark \\
DriveX~\cite{yang2024driving}        & \cmark & \cmark & \xmark  \\
FreeSim~\cite{fan2025freesim}          & \cmark & \cmark & \xmark \\
\midrule
DrivingForward~\cite{tian2025drivingforward}        & \cmark & \xmark  & \cmark \\
Omni-Scene~\cite{wei2025omni}          & \cmark & \xmark & \cmark \\
\midrule
FreeVS~\cite{wang2024freevs}        & \xmark & \cmark & \cmark \\
DiST-4D~\cite{guo2025dist}          & \xmark & \cmark & \cmark \\
\midrule
\rowcolor{gray!10}
FreeGen (Ours)              & \cmark & \cmark & \cmark \\
\bottomrule
\end{tabular}
}
\vspace{-2ex}
\label{tab:method_comparison}
\end{table}

To this end, a practical solution should satisfy three requirements:
(1) \textbf{Interpolation Consistency.} Preserve geometric consistency with the observed views in their overlapping regions.
(2) \textbf{Extrapolation Realism.} Generate visually plausible content that matches the underlying distribution at unseen viewpoints.
(3) \textbf{Feed-forward.} Synthesize the scene in a single forward pass instead of iterative refinement, enabling scalability with data-driven advances (e.g., scaling laws) for large-scale driving scene generation.

\begin{figure*}[ht!]
    \centering
    \includegraphics[width=1.0\linewidth]{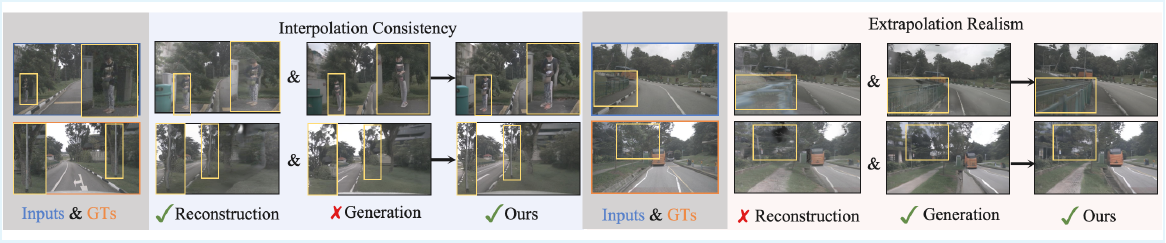}
    \vspace{-3ex}
    \caption{\textbf{Complementary strengths of reconstruction and generation methods.} Reconstruction maintains geometry but lacks realistic textures, generation improves realism but often distorts geometry. Our method combines both, achieving consistent and realistic results.}
    \label{fig:motivation}
    \vspace{-2ex}
\end{figure*}
We compare current methods in Tab.~\ref{tab:method_comparison}. Traditional reconstruction-based methods~\cite{yan2024street, chen2023periodic, yang2023emernerf, chen2024omnire} excel at enforcing interpolation consistency but are fundamentally limited to interpolating within sparse observations, making it difficult to render free-viewpoint views. They typically require per-scene optimization, leading to prohibitive computational costs that hinder large-scale scene synthesis. Recent feed-forward reconstruction approaches~\cite{tian2025drivingforward, wei2025omni} enable efficient scene reconstruction in a single pass but still struggle to produce realistic details at extrapolated views. With the rapid progress of generative models, data-driven methods trained on large-scale videos~\cite{blattmann2023stable, gao2024vista, wang2024drivedreamer} show strong potential for generating realistic views from sparse inputs, yet they generally lack fine-grained control over camera trajectories. Hybrid methods that combine generative models with reconstruction-based pipelines~\cite{zhao2025drivedreamer4d, ni2025recondreamer, yang2024driving, fan2025freesim} enhance visual detail while preserving interpolation consistency, but often require hours to days of per-scene training, undermining their practicality for feed-forward inference. 

Recent methods leverage free-viewpoint rendered views to guide generative refinement for high-quality feed-forward scene synthesis~\cite{wang2024freevs, guo2025dist}. However, LiDAR conditions~\cite{wang2024freevs} are sparse and expensive, making it difficult to cover distant buildings and sky, while image warping conditions~\cite{guo2025dist} often introduce structural misalignment (e.g., missing utility poles and distorted pedestrians in Fig.~\ref{fig:motivation}). Moreover, these conditions are non-learnable, limiting their ability to benefit from large-scale image data.

We observe that reconstruction and generation approaches offer complementary strengths for free-viewpoint driving scene synthesis. As shown in Fig.~\ref{fig:motivation}, reconstruction models excel at preserving geometric and interpolation consistency but are limited in fine details at unseen viewpoints. In contrast, generation models can produce visually realistic results from sparse observations, yet often introduce structural distortions with respect to the input views. This complementarity motivates us to unify these two paradigms in a single feed-forward framework that preserves geometric faithfulness while leveraging generative realism.

To address the aforementioned challenges, we propose \textbf{FreeGen}, a feed-forward reconstruction–generation framework for free-viewpoint driving scene synthesis.
We first employs a feed-forward 3D Gaussian Splatting (3DGS) model to efficiently reconstruct scene from sparse input images without per-scene optimization. Then We renders views along free-viewpoint trajectories to enforce interpolation consistency with the observed inputs. These rendered views are then used as geometric guidance for a generative refinement model, enabling realistic extrapolation at unseen viewpoints. Moreover, FreeGen does not require additional expensive annotations such as LiDAR or bounding boxes, relying solely on a single trajectory image data, which make it practical for large-scale driving scene synthesis.

Our method builds upon recent advances in feed-forward 3DGS~\cite{wei2025omni} and video diffusion models~\cite{blattmann2023stable}. To fully exploit their complementary strengths, we introduce two key designs in FreeGen. The reconstruction branch already produces structurally complete renderings with minimal visible holes, but due to limited supervision from a single trajectory, a naive generative module cannot reliably localize subtle artifacts. To address this, we \textbf{construct geometry conditions} based on the 3DGS rendering mechanism and incorporate a \textbf{geometry-aware diffusion refinement} module to guide appearance refinement. 
In addition, we propose a \textbf{co-training} strategy that samples viewpoints off the input trajectory and forms a closed loop between reconstruction and generation. The generative priors distilled during refinement are fed back into the reconstruction model to improve free-viewpoint rendering quality, while the reconstruction model provides geometric guidance to construct closed-loop supervision for the generative module.

Our contributions are summarized as follows:
\begin{itemize}
    \item We introduce FreeGen, a feed-forward reconstruction-generation co-training framework for free-viewpoint driving scene synthesis, which achieves both interpolation consistency and extrapolation realism without per-scene optimization or additional expensive annotations.
    \item We propose a geometry-aware diffusion refinement module to achieve high-quality refinement on reconstructed views. To further improve both reconstruction and generation performance at off-trajectory viewpoints, we introduce a closed-loop co-training strategy.
    \item Extensive experiments demonstrate that FreeGen outperforms existing approaches in free-viewpoint driving scene synthesis.
\end{itemize}

\section{Related Work}
\label{sec:related}

\begin{figure*}[!ht]
    \centering
    \includegraphics[width=1.0\linewidth]{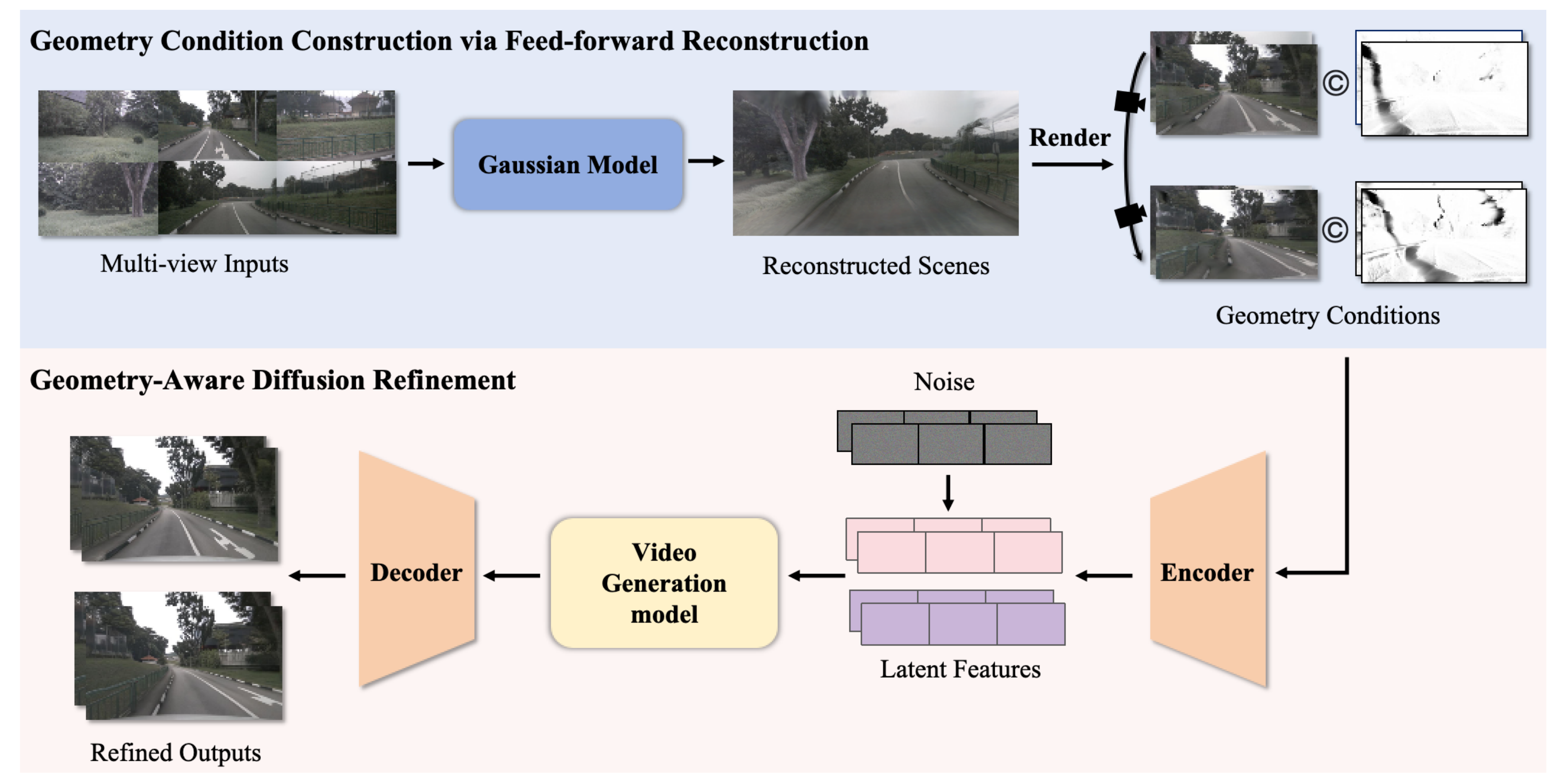}
    \vspace{-3ex}
    \caption{\textbf{Overview of the proposed FreeGen.} The reconstruction model encodes multi-view inputs into Gaussian features and decodes them into 3DGS representations. The rendered views and corresponding opacity maps are then used to guide the geometry-aware diffusion model, which performs fine-grained refinement. For clarity, depth maps are omitted from the illustration.}
    \label{fig:overview}
    \vspace{-3ex}
\end{figure*}

\noindent\textbf{Driving Scene Synthesis with Reconstruction.}
Early methods employed Neural Radiance Fields (NeRF)~\cite{barron2022mip,barron2023zip,mildenhall2021nerf,muller2022instant} to model scenes with MLPs and enable free-viewpoint image synthesis, but they suffer from slow training and rendering. The advent of 3DGS~\cite{kerbl20233d} has significantly advanced this paradigm. By leveraging explicit point-based representations and rasterization-based rendering, 3DGS achieves much faster training and inference. Recent works~\cite{wei2025omni,yan2024street} extend 3DGS to driving scenarios. However, they mainly focus on viewpoints along the training trajectory and often rely on per-scene optimization, which limits their scalability for large-scale driving scene synthesis. In addition, most of these methods~\cite{wei2025omni,yan2024street} require extra expensive annotations, such as point clouds or bounding boxes.
Recent feed-forward 3DGS~\cite{tian2025drivingforward,charatan2024pixelsplat,chen2024mvsplat} methods achieve efficient scene reconstruction by predicting 3D Gaussians in a single forward pass. Omni-Scene~\cite{wei2025omni} further combines pixel and voxel branches to improve render quality at viewpoints beyond the input views, leading to higher-quality driving scene reconstruction. However, these approaches still primarily target viewpoints along or near the training trajectory.

\noindent\textbf{Driving Scene Synthesis with Diffusion Prior.}
Thanks to the rapid development of diffusion-based generative methods~\cite{ho2020denoising,song2020denoising,dhariwal2021diffusion,ho2022classifier,rombach2022high,ho2022video}, image generation for driving scenes has made notable progress in recent years~\cite{gao2024vista,wang2024drivedreamer,gao2023magicdrive}. Many works apply diffusion models to free-viewpoint image synthesis in driving environments~\cite{blattmann2023stable}. Some approaches use diffusion models to refine free-viewpoint rendered views and iteratively optimize the scene representation. DriveDreamer4D~\cite{zhao2025drivedreamer4d} employs video diffusion models conditioned on initial recorded frames to synthesize free-viewpoint sequences and optimize dynamic scene representations, but it struggles to precisely control generated content and often suffers from structural misalignment. ReconDreamer~\cite{zhao2025drivedreamer4d} trains a video diffusion model on degraded renderings obtained during Gaussian Splatting training to repair free-viewpoint results. DriveX~\cite{yang2024driving} formulates an inverse problem, using diffusion models to refine free-viewpoint renderings before iteratively updating the scene representation. However, these methods require per-scene optimization during both training and inference, often taking hours to days for a single scene, which makes them difficult to scale to large-scale applications.
Other approaches avoid explicitly constructing full 3D scene representations and instead rely on LiDAR or image warping as free-viewpoint rendering proxies. FreeVS~\cite{wang2024freevs} guides diffusion-based generation by aggregating multi-frame LiDAR projections, enabling efficient training and feed-forward inference. 
However, LiDAR inputs are often costly and spatially sparse, especially for distant objects and sky regions, which leads to incomplete guidance. 
DiST-4D~\cite{guo2025dist} adopts image warping to project observed views to free-viewpoints for diffusion-based generation, but such warping is prone to structural misalignment, making it difficult to maintain consistency between the results and the observations.
\section{Methodology}
\label{sec:methods}
\begin{figure*}[t!]
    \centering
    \includegraphics[width=1.0\linewidth]{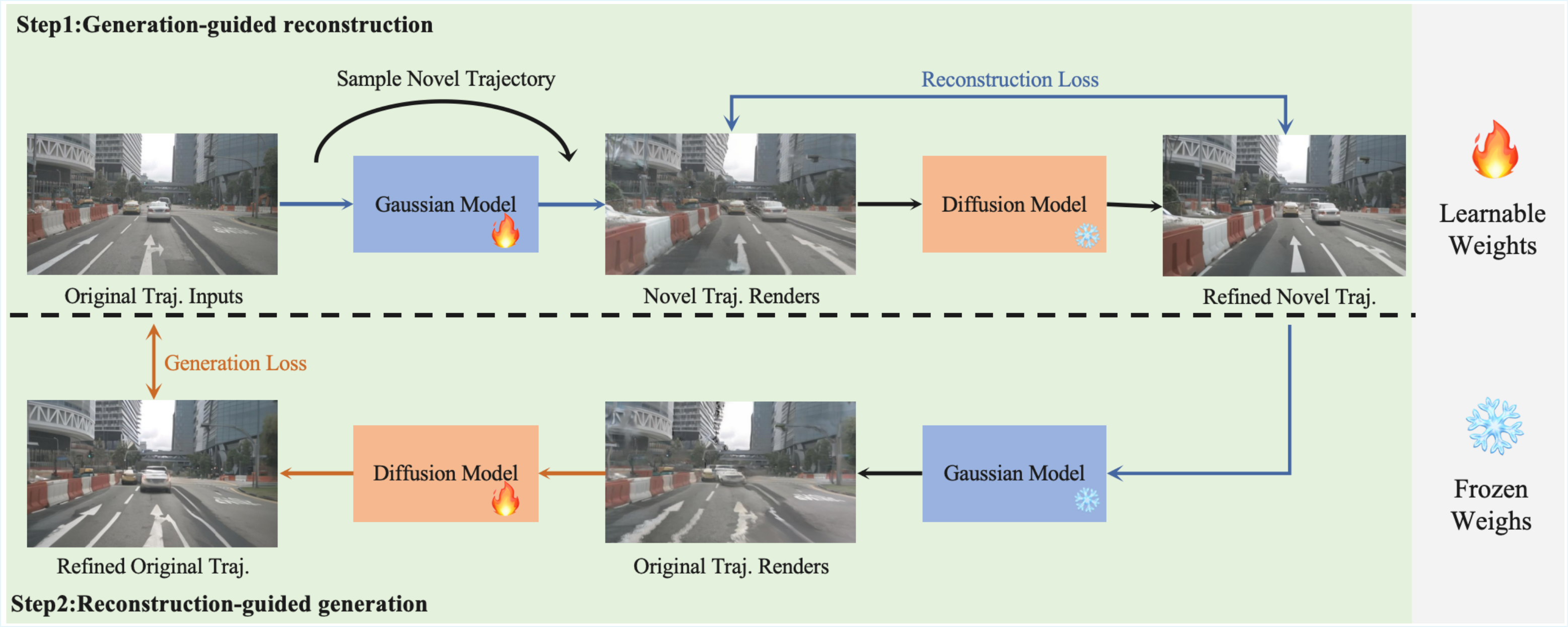}
    \vspace{-1ex}
    \caption{\textbf{Illustration of the Co-Training strategy.} FreeGen adopts a closed-loop co-training strategy between the Gaussian reconstruction model and the diffusion refinement model. Novel trajectories~(Traj.) are sampled and rendered by the Gaussian model, then refined by the diffusion model to form pseudo-supervision. The refined results are fed back to the Gaussian model for reconstruction loss, while the diffusion model learns from generation loss on the original trajectory. }
    \label{fig:cot}
    \vspace{-1ex}
\end{figure*}

Our key idea is to leverage the complementary strengths of reconstruction and generation to enhance both interpolation consistency and extrapolation realism in free-viewpoint scene synthesis. To achieve this, we first reconstruct the scene from multi-view inputs using a feed-forward reconstruction model, following~\cite{wei2025omni}. Then we perform free-viewpoint rendering along given camera trajectories to obtain rendered views and corresponding geometry conditions (Sec.~\ref{sec:feed_forward_reconstruction}). Next, the rendered results are used to guide a diffusion refinement model (Sec.~\ref{sec:geometry_aware_diffusion_refinement}). The overall framework of our method is illustrated in Fig.~\ref{fig:overview}. Furthermore, we jointly train the reconstruction and generation models to improve overall performance (Sec.~\ref{sec:cotraining}), as shown in Fig.~\ref{fig:cot}.

\begin{table*}[!ht]
\caption{ \textbf{Quantitative comparison of free-viewpoint synthesis quality.} We compare FreeGen with existing methods under different trajectory shift settings.
For fairness, the results of all baseline methods are taken from~\cite{guo2025dist}.
Auxiliary Information (Aux. Info.) indicates whether a method relies on additional data beyond images during training or inference, such as LiDAR point clouds or 3D bounding box annotations.
All models are evaluated on six camera views.}
\vspace{-1ex}
\centering
\small
\setlength{\tabcolsep}{15pt}
\begin{tabular}{c | c | cc | cc | cc}
\toprule
\multirow{2}{*}{\textbf{Method}} & \multirow{2}{*}{\textbf{Aux. Info.}} &
\multicolumn{2}{c|}{\textbf{Shift $\pm$1m}} &
\multicolumn{2}{c|}{\textbf{Shift $\pm$2m}} &
\multicolumn{2}{c}{\textbf{Shift $\pm$4m}} \\
& &
FID$\downarrow$ & FVD$\downarrow$ &
FID$\downarrow$ & FVD$\downarrow$ &
FID$\downarrow$ & FVD$\downarrow$ \\
\midrule
PVG~\cite{chen2023periodic}              & LiDAR \& Box & 48.15 & 246.74 & 60.44 & 356.23 & 84.50 & 501.16 \\
EmerNeRF~\cite{yang2023emernerf}    & LiDAR \& Box     & 37.57 & 171.47 & 52.03 & 294.55 & 76.11 & 497.85 \\
StreetGaussian~\cite{yan2024street}
                            & LiDAR \& Box     & 32.12 & 153.45 & 43.24 & 256.91 & 67.44 & 429.98 \\
OmniRe~\cite{chen2024omnire}        & LiDAR \& Box & 31.48 & 152.01 & 43.31 & 254.52 & 67.36 & 428.20 \\
FreeVS~\cite{wang2024freevs}       & LiDAR \& Box        & 51.26 & 431.99 & 62.04 & 497.37 & 77.14 & 556.14 \\
DiST-4D~\cite{guo2025dist}             & LiDAR \& Box        & 10.12 & 45.14 & 12.97 & 68.80 & 17.57 & 105.29 \\
\midrule
\rowcolor{gray!10}
    FreeGen (Ours)              & None        & \textbf{9.49} & \textbf{32.83} & \textbf{11.34} & \textbf{44.98} & \textbf{14.44} & \textbf{70.75} \\
\bottomrule
\end{tabular}
\label{tab:shift_comparison}
\vspace{-1ex}
\end{table*}

\begin{figure*}[!ht]
    \centering
    \includegraphics[width=1.0\linewidth]{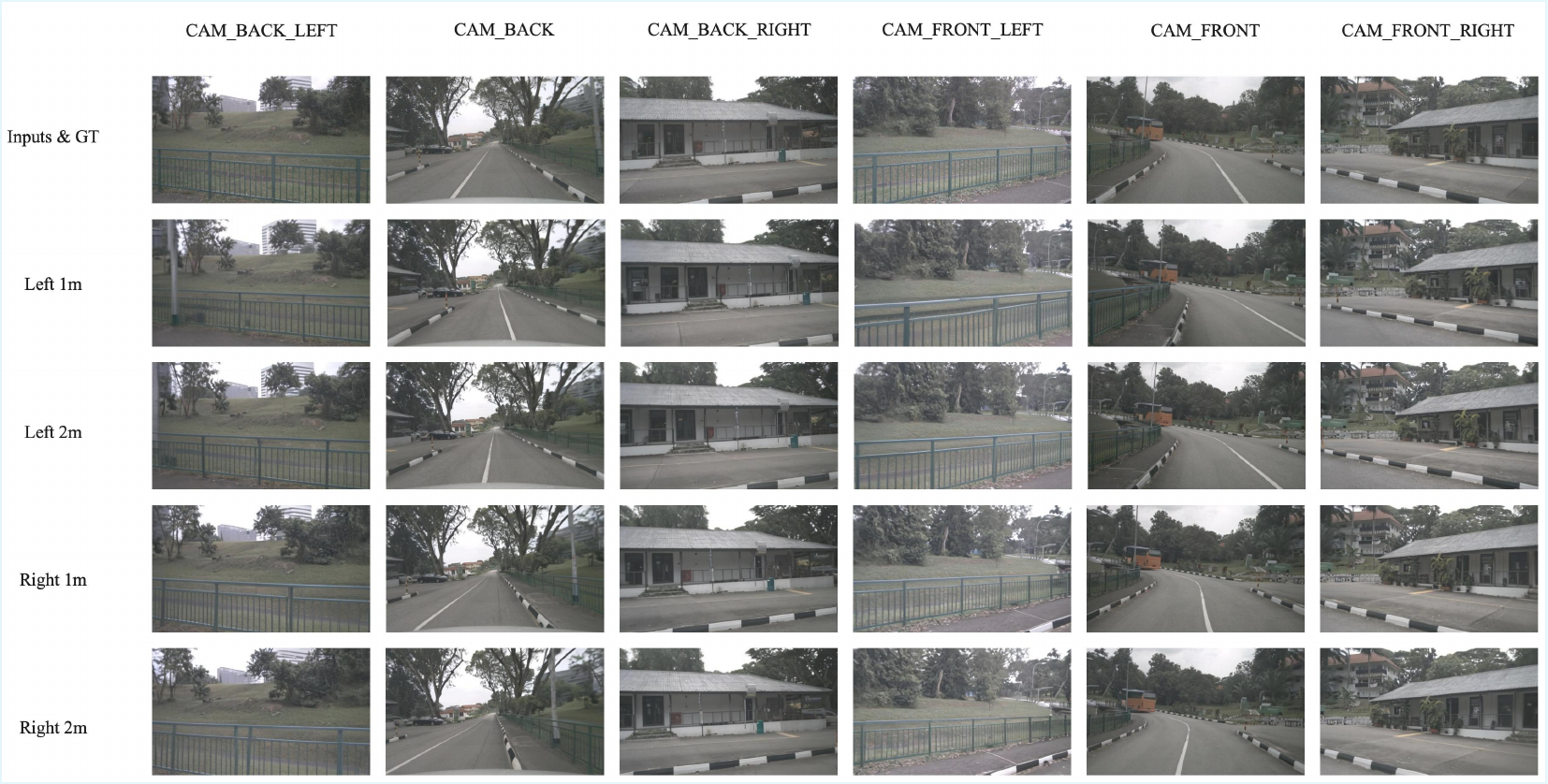}
    \vspace{-2ex}
    \caption{\textbf{Qualitative Results under Spatial Viewpoint Shifts.} We show generation results under lateral shifts of \(1\text{m}\) and \(2\text{m}\) to the left and right with six input views. Our method generates high-quality and detailed scenes while maintaining consistency across different viewpoints.}
    \label{fig:res}
    \vspace{-2ex}
\end{figure*}

\begin{figure}[!ht]
    \centering
    \includegraphics[width=1.0\linewidth]{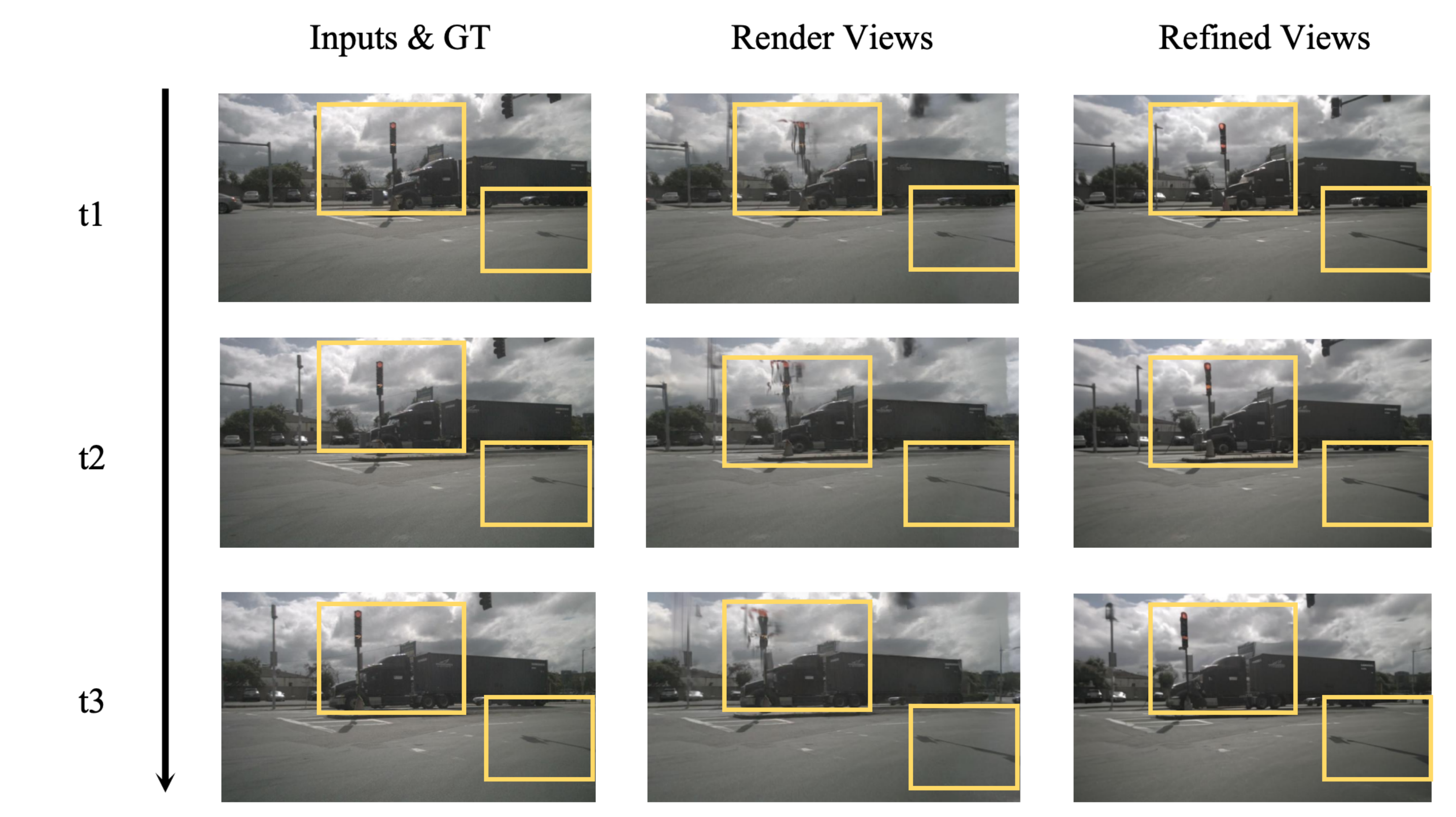}
    \vspace{-2ex}
    \caption{\textbf{Qualitative Results of Temporal Consistency.} We show three consecutive frames synthesized under a $1\text{m}$ lateral shift. For each timestamp, we present the ground-truth recorded image, the rendered views from the reconstruction model, and the refined results. The refined sequences remain temporally stable and preserve object geometry across time, highlighting the strong temporal consistency achieved by our method.
}
    \label{fig:time}
    \vspace{-2ex}
\end{figure}

\begin{table}[!ht]
\caption{
\textbf{Quantitative comparison on recorded views across different metrics.} We evaluate the performance on the original trajectory. \textbf{Image warping} refers to the geometric condition used by DiST-4D, while \textbf{Our Rec. model} is our reconstruction model.
}
\vspace{-1ex}
\centering
\small
\setlength{\tabcolsep}{11pt}
\begin{tabular}{cccccc}
\toprule
\textbf{Method} & \textbf{PSNR}↑ & \textbf{SSIM}↑ & \textbf{LPIPS}↓ \\
\midrule
Image warping       & 18.14 & 0.58 & 0.42 \\
Our Rec. model      & 20.40 & 0.62 & 0.36   \\
\midrule
DiST-4D~\cite{guo2025dist}
  & 23.85  & 0.74 & \textbf{0.29}\\
\rowcolor{gray!10}
FreeGen (Ours) & \textbf{24.19} & \textbf{0.75} & \textbf{0.29} \\
\bottomrule
\end{tabular}
\label{tab:recorded_views}
\vspace{-1ex}
\end{table}

\subsection{Geometry Condition Construction via Feed-forward Reconstruction}
\label{sec:feed_forward_reconstruction}
An appropriate scene representation should support both efficient reconstruction and fast rendering. Recent feed-forward 3DGS methods~\cite{charatan2024pixelsplat,chen2024mvsplat,wei2025omni,tian2025drivingforward} make this possible by reconstructing large-scale driving scenes from multi-view images within seconds, achieving real-time rendering and learning rich scene priors from large-scale data. Motivated by this, we adopt a feed-forward 3DGS-based reconstruction network in FreeGen, following~\cite{wei2025omni}.

Given sparse $N$-view inputs $\{I_i\}_{i=1}^N$ with corresponding camera parameters $\{P_i\}_{i=1}^N$, we first predict depth maps $\{D_i\}_{i=1}^N$ using an off-the-shelf depth estimation model~\cite{hu2024metric3d} to assist scene reconstruction. We then use a Gaussian encoder to extract Gaussian features. Specifically, we apply a multi-view U-Net style pixel encoder~\cite{charatan2024pixelsplat,chen2024mvsplat,tian2025drivingforward,wei2025omni} to extract pixel-aligned high-resolution feature maps $\{\mathcal{F}^{\text{pixel}}_i\}_{i=1}^N$. Then we fuse multi-view features into a triplane representation via a Triplane Transformer~\cite{huang2023tri} with deformable attention~\cite{li2024bevformer}.
Finally, Gaussian decoders transform both pixel and voxel features into a set of $K$ 3D Gaussians~\cite{kerbl20233d}
$\{G_i = (\delta_i, \alpha_i, s_i, q_i, c_i)\}_{i=1}^K$, where $\delta_i$, $\alpha_i$, $s_i$, $q_i$, and $c_i$ denote the position, opacity, scale, orientation, and color of each Gaussian.
We then render the color image \(I_{\mathrm{geo}}\), opacity map \(A_{\mathrm{geo}}\), and depth map \(D_{\mathrm{geo}}\) via tile-based rasterization~\cite{kerbl20233d}. 
The color image is computed as follows:
\begin{equation}
I_{\mathrm{geo}}
= \sum_{i=1}^{K} c_i \,\alpha'_i\,
  \prod_{j=1}^{i-1}\!\bigl(1 - \alpha'_j\bigr),
\end{equation}
where \(\alpha'_i\) is the screen-space opacity contribution of the \(i\)-th Gaussian determined as 3DGS~\cite{kerbl20233d}.
The opacity map \(A_{\mathrm{geo}}\) and depth map \(D_{\mathrm{geo}}\) are obtained similar to the color image. Then we define the geometry conditions \(C_{\text{geo}}\) as follows:
\begin{equation}
C_{\mathrm{geo}} \;=\;
\bigl[\, I_{\mathrm{geo}},\, D_{\mathrm{geo}},\, A_{\mathrm{geo}} \,\bigr].
\label{eq:triplet}
\end{equation}

In FreeGen, the geometry conditions serve as the geometric condition for the diffusion-based generation model, enabling geometry-aware refinement.
The color image provides structural and appearance information, the depth map enhances geometric consistency~\cite{guo2025dist}, and the opacity map reflects the reliability of the color image, where higher opacity usually corresponds to more reliable renderings.

The training phase of the reconstruction model includes two stages. In the first stage, we follow~\cite{wei2025omni} and train the model on the original single trajectory sequences, forming cross-timestamp pairs to establish reconstruction capability of the model. 
In the second training stage, we randomly render views at off-trajectory viewpoints and refine them with the generation model to obtain pseudo-labels. These pseudo-labels then supervise the reconstruction model to improve free-viewpoint rendering. Details are provided in Sec.~\ref{sec:cotraining}.
In the overall training phase, 
We supervise the reconstruction model using mean squared error loss, LPIPS loss~\cite{zhang2018unreasonable}, and L1 depth loss as follows:
\begin{equation}
\mathcal{L}_{recon} = \mathcal{L}_{mse} + \lambda_{1}\mathcal{L}_{lpips} + \lambda_{2}\mathcal{L}_{depth},
\end{equation}
where $\lambda_{1}$ and $\lambda_{2}$ are the weights for each loss component. The depth supervision uses pseudo-labels produced by the off-the-shelf depth model~\cite{hu2024metric3d}.

\subsection{Geometry-Aware Diffusion Refinement}
\label{sec:geometry_aware_diffusion_refinement}

Although the reconstruction model can effectively recover scene structures from multi-view images, it still struggles to produce high-quality and realistic results at free-viewpoint, especially for off-trajectory views.
With the recent success of diffusion models in image generation~\cite{ho2020denoising,rombach2022high}, we introduce a diffusion-based generation module to refine the feed-forward 3DGS renderings.
Since the video diffusion model~\cite{blattmann2023stable} has been extensively pre-trained on large-scale data, introducing additional parameters may degrade its generalization ability. Therefore, we keep our modification minimal to maintain compatibility and ensure stable generalization across diverse scenes.

To simplify the explanation, we describe the process using single image as examples. In practice, we employ a video diffusion model to ensure temporal consistency.
For the geometry conditions \(C_{\mathrm{geo}}\) rendered by the reconstruction model, 
we encode the geometry conditions into a condition latent representation \(\mathbf{z}_c = \mathcal{E}_c(C_{\mathrm{geo}})\) through a learnable condition encoder \(\mathcal{E}_c\) like~\cite{guo2025dist}.
During training, we encode the reference image \(I_r\) using a frozen image encoder \(\mathcal{E}_v\) to obtain the latent representation \(\mathbf{z}_v = \mathcal{E}_v(I_r)\).  
Gaussian noise is added to the latent to obtain a noisy version~\cite{ho2020denoising} as follows:
\begin{equation}
\mathbf{z}_\tau = \alpha_\tau \mathbf{z}_v + \sigma_\tau \boldsymbol{\epsilon},
\end{equation}
where \(\boldsymbol{\epsilon} \sim \mathcal{N}(0,1)\) and \(\tau\) is the diffusion timestep.
The geometry condition latent \(\mathbf{z}_c\) is concatenated with the noise latent \(\mathbf{z}_\tau\) along the channel dimension and fed into the diffusion model \(f_\theta\) to predict the noise as follows:
\begin{equation}
\hat{\boldsymbol{\epsilon}} = f_\theta(\mathbf{z}_\tau, \mathbf{z}_c).
\end{equation}

The training objective~\cite{ho2020denoising} minimizes the mean squared error between the predicted and true noise as follows:
\begin{equation}
\mathcal{L}_{\mathrm{gen}} = \mathbb{E}_{\tau, \mathbf{z}_v, \boldsymbol{\epsilon}}\big[\|\boldsymbol{\epsilon} - f_\theta(\mathbf{z}_\tau, \mathbf{z}_c)\|_2^2\big].
\end{equation}
This allows the diffusion model to learn geometry-aware refinement conditioned on the rendered scene structure.

The generation model is also trained in two stages.
In the first stage, we supervise the model with renderings and the corresponding ground–truth frames so that it learns to refine effectively under geometric conditions.
In the second stage, we introduce a closed–loop supervision scheme to improve generation quality at off–trajectory viewpoints.
Details are provided in Sec.~\ref{sec:cotraining}.
The depth supervision follows the same protocol in Sec.~\ref{sec:feed_forward_reconstruction}.

\subsection{Co-Training}
\label{sec:cotraining}

Considering that both the reconstruction and generation models are trained with the single trajectory data, coverage of viewpoint variation is limited for free-viewpoint synthesis.
Inspired by closed-loop training~\cite{guo2025dist, ren2025gen3c} and noting that both models are learnable, we design a co-training scheme that lets the two models promote each other, as shown in Fig.~\ref{fig:cot}.
The procedure alternates two steps and freezes one model at a time to keep optimization stable.

\noindent\textbf{Step 1: Generation-guided reconstruction.}
We randomly sample off-trajectory viewpoints and render them with the Gaussian model to obtain free-viewpoint images and their geometry conditions.
The frozen diffusion model refines these off-trajectory renders and produces pseudo-labels.
The Gaussian model is then updated on these pseudo-labels using the reconstruction losses defined in Sec.~\ref{sec:feed_forward_reconstruction}.
This step transfers appearance priors to the reconstruction branch and improves its free-viewpoint rendering.

\noindent\textbf{Step 2: Reconstruction-guided generation.}
Using the updated Gaussian model, we take the off-trajectory renders and their refined results from Step 1 and render them back to the original trajectory viewpoints.
The diffusion model refines these re-rendered views under geometric conditions and the outputs are supervised by the ground-truth frames with the generation losses in Sec.~\ref{sec:geometry_aware_diffusion_refinement}.
This step strengthens geometry-aware refinement in the generation branch.

Through alternating Step 1 and Step 2 the reconstruction branch learns stronger off-trajectory geometry and the generation branch learns geometry-consistent appearance.
The closed loop progressively aligns structural consistency and visual realism across free-viewpoints.
\section{Experiments}
\label{sec:exp}

\begin{figure*}[!ht]
    \centering
    \includegraphics[width=1.0\linewidth]{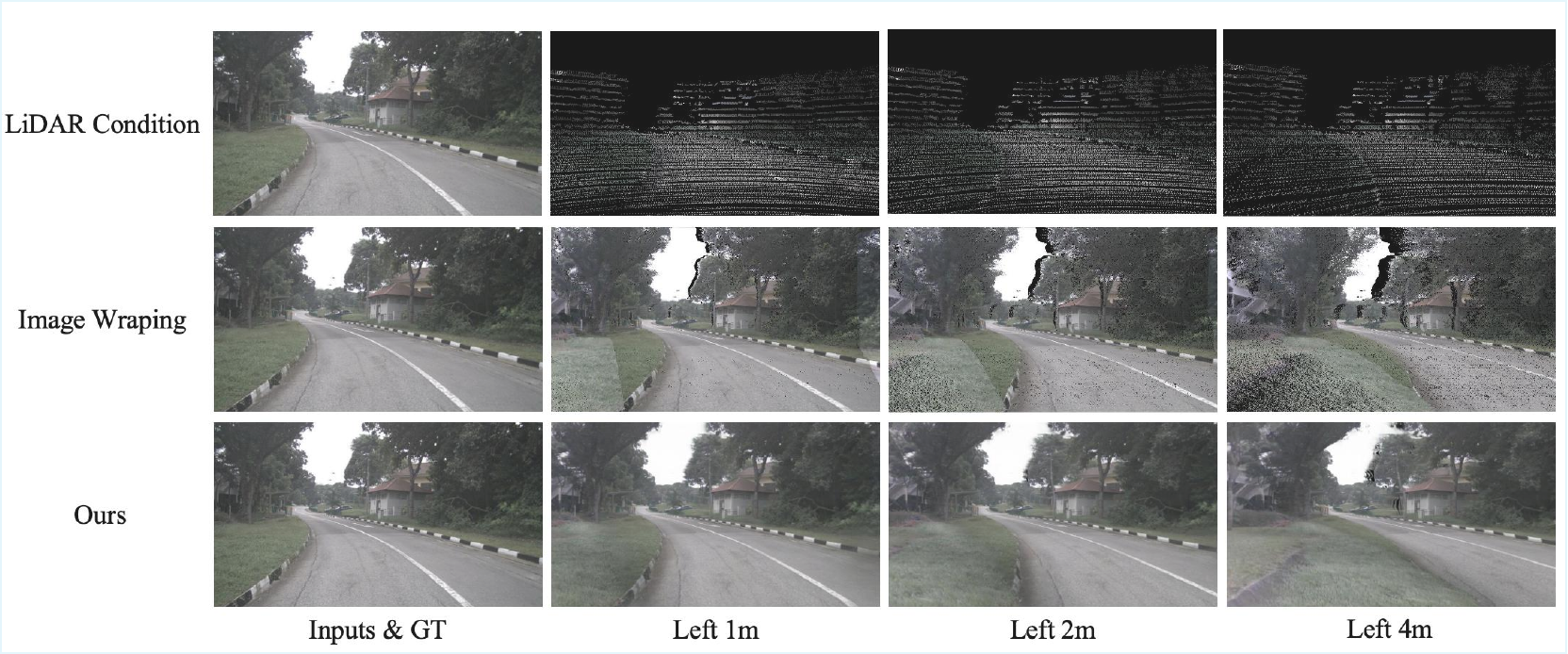}
    \vspace{-2ex}
    \caption{
    \textbf{Qualitative comparison of geometric conditions.}
    We construct target views with lateral shifts of 1\,m, 2\,m, and 4\,m.
    For LiDAR-based conditions, we aggregate 10 consecutive frames and project the point clouds to the target view.
    For image warping, we use depth to warp the input images to the target camera.
    Our condition is obtained by rendering from the feed-forward reconstruction model.
    }
    \label{fig:cond}
    \vspace{-2ex}
\end{figure*}

\subsection{Setup}
\noindent\textbf{Dataset.}
We conduct experiments on nuScenes~\cite{caesar2020nuscenes}.
Following prior work~\cite{guo2025dist,wei2025omni}, the 2\,Hz keyframe annotations are interpolated to 12\,Hz.
We use the official split with 700 scenes for training and 150 scenes for validation.

\noindent\textbf{Evaluation.}
For \textit{off-trajectory evaluation}, we follow the protocols in~\cite{guo2025dist, wang2024freevs}. 
Specifically, we uniformly sample frames every two frames along the original recorded trajectory, and apply lateral camera shifts 
$\tau \in \{\pm1\text{m}, \pm2\text{m}, \pm4\text{m}\}$ to each sampled frame.
We synthesize RGB videos under the shifted viewpoints and compute Fréchet Inception Distance (FID)~\cite{Seitzer2020FID} and Fréchet Video Distance (FVD)~\cite{stylegan_v} between the synthesized videos and the ground-truth videos rendered from the original trajectory. These metrics reflect appearance realism and temporal coherence under viewpoint extrapolation.
For \textit{on-trajectory evaluation}, we also sample frames every two frames, consistent with the training setup in~\cite{guo2025dist}, and generate views by shifting each sampled camera pose toward the next sampled viewpoint. 
We report peak signal-to-noise ratio (PSNR), structural similarity index (SSIM)~\cite{wang2004image}, and perceptual similarity 
(LPIPS)~\cite{zhang2018unreasonable} to assess reconstruction accuracy and perceptual fidelity along the original trajectory.

\noindent\textbf{Implementation details.}
We use  \(768\times432\) resolution with video length of \(T=6\) for both training and inference, following \cite{guo2025dist}.
The video diffusion model is initialized from Stable Video Diffusion~(SVD)~\cite{blattmann2023stable}, following~\cite{guo2025dist,wang2024freevs}.
$\lambda_{1}$ and $\lambda_{2}$ are set to \(0.05\) and \(0.01\) respectively, following~\cite{wei2025omni}.
All experiments run on two NVIDIA A100 GPUs.
Please refer to the supplements for more implementation details.

\begin{table}[!t]
\caption{
\textbf{Ablation study on the contribution of each module.}
All metrics are evaluated under left \(2\text{m}\) shift using six camera views. \textbf{Rec.} and \textbf{Gen.} indicate enabling the reconstruction and generation branches.
\textbf{Opac.} denotes using the opacity maps as additional geometric condition.
\textbf{Rec. CT} and \textbf{Gen. CT} indicate co-training the reconstruction branch and the generation branch respectively.
}
\vspace{-1ex}
\centering
\setlength{\tabcolsep}{4pt}
\resizebox{\linewidth}{!}{
\begin{tabular}{c | ccccc | cc}
\toprule
\textbf{Settings} & \textbf{Rec.} & \textbf{Gen.} & \textbf{Opac.} & \textbf{Rec. CT} & \textbf{Gen. CT} & \textbf{FID-2m}$\downarrow$ & \textbf{FVD-2m}$\downarrow$ \\
\midrule
(a) & \checkmark &  &  &  &  & 23.25 & 246.73 \\
(b) & \checkmark & \checkmark &  &  &  & 15.49 & 60.53 \\
(c) & \checkmark & \checkmark & \checkmark &  &  & 13.86 & 57.21 \\
(d) & \checkmark & \checkmark & \checkmark & \checkmark &  & 14.39 & 54.93 \\
(e) & \checkmark & \checkmark & \checkmark &  & \checkmark & 13.16 & 51.78 \\
(f) & \checkmark & \checkmark & \checkmark & \checkmark & \checkmark & \textbf{13.05} & \textbf{50.92} \\
\bottomrule
\end{tabular}
}
\label{tab:ablation_modules}
\vspace{-2ex}
\end{table}

\subsection{Main Results}

\noindent\textbf{Off-trajectory Quantitative Evaluation.}
Table~\ref{tab:shift_comparison} reports results under three lateral shift settings.
Classical reconstruction methods~\cite{chen2023periodic,yang2023emernerf,yan2024street,chen2024omnire} suffer from blur at off-trajectory viewpoints because no observations exist outside the input views, and they usually rely on LiDAR and boxes to supply depth and category cues for reconstruction.
FreeVS~\cite{wang2024freevs} introduces extra geometry conditions by projecting LiDAR, but the point cloud is sparse and the aggregation requires boxes to separate dynamic from static objects, which limits performance.
DiST-4D~\cite{guo2025dist} uses image warping to provide geometry conditions and obtains noticeable gains, yet warping cannot faithfully represent full scene structure and the pipeline still depends on LiDAR and boxes with heavy preprocessing to recover depth.
In contrast, our FreeGen achieves the best results at all shift levels.
The improvements are especially large on FVD, indicating higher temporal coherence and realism under free-viewpoint changes.
FreeGen uses only images and an off-the-shelf depth estimator to obtain depth in seconds, which is fast and easy to deploy at large scale.

\noindent\textbf{On-trajectory Quantitative Evaluation.}
Table~\ref{tab:recorded_views} presents the results on the original recorded trajectory.
We first compare two types of geometric conditions on the recorded trajectory, namely the image warping conditions used in DiST-4D~\cite{guo2025dist} and the geometry conditions rendered by our feed forward reconstruction model.
As shown in Table~\ref{tab:recorded_views}, our reconstruction model provides more accurate geometry and produces images that are closer to the ground truth in both structure and perceptual quality.
Table~\ref{tab:recorded_views} also reports the results after generative refinement.
FreeGen slightly improves over DiST-4D~\cite{guo2025dist} on PSNR and SSIM and matches its LPIPS.
DiST-4D benefits from high quality depth maps that are constructed on the original trajectory using LiDAR and handcrafted processing.
In contrast, our FreeGen only relies on an off-the-shelf depth model~\cite{hu2024metric3d} yet still achieves comparable or better reconstruction quality on recorded trajectory views and stronger temporal and perceptual performance on off-trajectory evaluations.
Moreover, when we compare the reconstructed views with the corresponding refined results, we observe that our co-training framework further enhances realism while still preserving high fidelity to the original scenes.

\noindent\textbf{Qualitative Results.}
Fig.~\ref{fig:res} shows qualitative results of FreeGen with six input views under lateral shifts of \(1\text{m}\) and \(2\text{m}\) to the left and right.
Our method synthesizes sharp and detailed appearances while preserving scene geometry across viewpoints.
Thin structures such as poles and lane markings remain intact and occlusion boundaries are stable.
Cross-view consistency is maintained under all shifts, indicating that the geometry-aware refinement effectively guides high-quality free-viewpoint generation. Fig.~\ref{fig:time} further presents qualitative results on temporal consistency.
Across consecutive frames the refined views remain stable, without noticeable flickering or geometry drift. 

\begin{figure}[!t]
    \centering
    \includegraphics[width=1.0\linewidth]{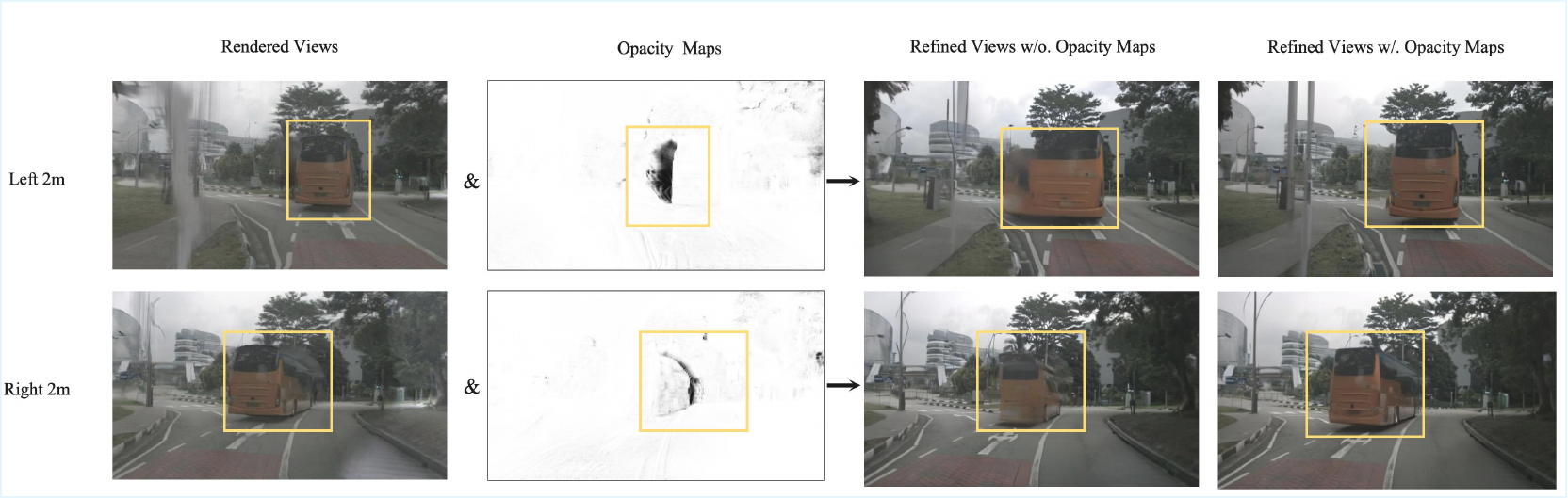}
    \vspace{-3ex}
    \caption{\textbf{Ablation on the opacity map.} We compare generation results with and without opacity guidance to assess its effect.}
    \label{fig:alpha}
    \vspace{-2ex}
\end{figure}
\subsection{Ablation Study}

\noindent\textbf{Main Ablation Studies.}
We conduct an ablation under a leftward shift of \( -2\text{m} \) to validate each design choice. Table~\ref{tab:ablation_modules} reports the results for settings (a) to (f).
\textit{(a) Reconstruction only~(Rec.).}
Novel views are rendered by the reconstruction model without generative refinement.
This yields competitive FID due to faithful geometry but limited off-trajectory information and single-frame input lead to worse FVD.
\textit{(b) Generation~(Gen.).}
Using the rendered color and depth to guide diffusion model improves both metrics, but the model lacks an explicit notion of reliable regions to refine.
\textit{(c) Opacity guidance~(Opac.).}
Adding the opacity map provides reliability cues, which further improves performance.
\textit{(d) Co-training on reconstruction only~(Rec.CT).}
We update the Gaussian branch with pseudo-labels refined by the generation model while keeping its trained on the single trajectory data.
FVD improves due to better temporal stability while FID slightly drops, reflecting limited pseudo-label quality.
\textit{(e) Co-training on generation only~(Gen.CT).}
We update the generation model with diverse off-trajectory renders.
Both FID and FVD improve since the generation model observes a wider distribution of viewpoints.
\textit{(f) Co-training on both models.}
Jointly updating reconstruction and generation models achieves the best results.
The closed loop transfers appearance priors to the Gaussian model and supplies geometry-consistent supervision to the generation model, leading to the highest quality free-viewpoint driving scene synthesis.

\noindent\textbf{Analysis of Geometry Conditions.}
We study how different geometry conditions affect generative refinement with both visualization and quantitative evaluation.
Fig.~\ref{fig:cond} compares three types of conditions.
FreeVS~\cite{wang2024freevs} projects LiDAR to the image plane.
Even after aggregating 10 adjacent frames, the LiDAR condition remains sparse and is almost unusable for distant structures.
DiST-4D~\cite{guo2025dist} warps images to the target view and shows clearer layouts than LiDAR, yet it cannot represent full scene geometry and exhibits noticeable structural drift at larger viewpoint shifts.
Our 3DGS-based condition captures scene structure and fine details in a few seconds and aligns well with the target view.
Table~\ref{tab:fid_fvd_comparison} reports the metrics for LiDAR and reconstruction conditions.
We exclude LiDAR projection from the table because the condition is too sparse to provide a fair comparison.
Although our feed-forward reconstruction model uses only multi-view inputs from a single timestep at inference, the rendered geometry for new trajectories outperforms the current best iterative reconstruction pipeline.
We attribute this to pretraining on large-scale data, which provides strong priors that improve predictions at off-trajectory viewpoints.

\noindent\textbf{Analysis of Opacity Guidance.}
In Fig.~\ref{fig:alpha}, we visualize the effect of opacity guidance.
The opacity map clearly highlights reliable and unreliable regions.
With this guidance, the generative model can focus refinement on uncertain areas, which improves the overall quality of the  results.

\begin{table}[!t]
\caption{\textbf{Quantitative evaluation of different geometry conditions.} We compare the image warping condition used in DiST-4D with our geometric condition rendered by the feed-forward reconstruction model under different viewpoint shifts, using FID and FVD as metrics.
OmniRe is included as a reference for the performance of the current best iterative reconstruction method.
}
\vspace{-1ex}
\centering
\setlength{\tabcolsep}{6pt}
\resizebox{\linewidth}{!}{
\begin{tabular}{c | cc | cc | cc}
\toprule
\multirow{2}{*}{\textbf{Method}} &
\multicolumn{2}{c|}{\textbf{Shift $\pm$1m}} &
\multicolumn{2}{c|}{\textbf{Shift $\pm$2m}} &
\multicolumn{2}{c}{\textbf{Shift $\pm$4m}} \\
& FID$\downarrow$ & FVD$\downarrow$ & 
FID$\downarrow$ & FVD$\downarrow$ & 
FID$\downarrow$ & FVD$\downarrow$ \\
\midrule
OmniRe         & 31.48 & 152.01 & 43.31 & 254.52 & 67.36 & 428.20 \\
\midrule
Image wraping & 32.27 & 216.69 & 60.64 & 344.68 & 93.27 & 398.36 \\
Our Rec. model       & \textbf{12.92} & \textbf{117.49} & \textbf{20.96} & \textbf{205.99} & \textbf{39.47} & \textbf{370.79} \\
\bottomrule
\end{tabular}
}
\label{tab:fid_fvd_comparison}
\vspace{-1ex}
\end{table}
\section{Conclusion}
\label{sec:conclu}

We have introduced FreeGen, a novel feed-forward framework for free-viewpoint driving scene synthesis from the single trajectory observations. Our approach effectively addresses the critical challenge of balancing interpolation consistency and extrapolation realism. By co-training a reconstruction model with a generation model, FreeGen leverages the stable geometric guidance from reconstruction to ensure structural consistency, while harnessing the power of generative priors to enhance visual realism at unseen viewpoints. Extensive experiments demonstrate that FreeGen surpasses existing methods, enabling the efficient synthesis of high-fidelity and consistent driving scenes. 

{
    \small
    \bibliographystyle{ieeenat_fullname}
    \bibliography{main}

@String(TOG= {ACM Trans. Graph.})

@String(AAAI = {AAAI})

@String(TOG   = {ACM TOG})

@inproceedings{hu2022st,
  title={St-p3: End-to-end vision-based autonomous driving via spatial-temporal feature learning},
  author={Hu, Shengchao and Chen, Li and Wu, Penghao and Li, Hongyang and Yan, Junchi and Tao, Dacheng},
  booktitle={European Conference on Computer Vision},
  pages={533--549},
  year={2022},
  organization={Springer}
}

@inproceedings{hu2023planning,
  title={Planning-oriented autonomous driving},
  author={Hu, Yihan and Yang, Jiazhi and Chen, Li and Li, Keyu and Sima, Chonghao and Zhu, Xizhou and Chai, Siqi and Du, Senyao and Lin, Tianwei and Wang, Wenhai and others},
  booktitle={Proceedings of the IEEE/CVF conference on computer vision and pattern recognition},
  pages={17853--17862},
  year={2023}
}

@inproceedings{jiang2023vad,
  title={Vad: Vectorized scene representation for efficient autonomous driving},
  author={Jiang, Bo and Chen, Shaoyu and Xu, Qing and Liao, Bencheng and Chen, Jiajie and Zhou, Helong and Zhang, Qian and Liu, Wenyu and Huang, Chang and Wang, Xinggang},
  booktitle={Proceedings of the IEEE/CVF International Conference on Computer Vision},
  pages={8340--8350},
  year={2023}
}

@inproceedings{caesar2020nuscenes,
  title={nuscenes: A multimodal dataset for autonomous driving},
  author={Caesar, Holger and Bankiti, Varun and Lang, Alex H and Vora, Sourabh and Liong, Venice Erin and Xu, Qiang and Krishnan, Anush and Pan, Yu and Baldan, Giancarlo and Beijbom, Oscar},
  booktitle={Proceedings of the IEEE/CVF conference on computer vision and pattern recognition},
  pages={11621--11631},
  year={2020}
}

@article{gao2023magicdrive,
  title={Magicdrive: Street view generation with diverse 3d geometry control},
  author={Gao, Ruiyuan and Chen, Kai and Xie, Enze and Hong, Lanqing and Li, Zhenguo and Yeung, Dit-Yan and Xu, Qiang},
  journal={arXiv preprint arXiv:2310.02601},
  year={2023}
}

@inproceedings{gao2025magicdrive,
  title={MagicDrive-V2: High-resolution long video generation for autonomous driving with adaptive control},
  author={Gao, Ruiyuan and Chen, Kai and Xiao, Bo and Hong, Lanqing and Li, Zhenguo and Xu, Qiang},
  booktitle={Proceedings of the IEEE/CVF International Conference on Computer Vision},
  pages={28135--28144},
  year={2025}
}

@inproceedings{yan2024street,
  title={Street gaussians: Modeling dynamic urban scenes with gaussian splatting},
  author={Yan, Yunzhi and Lin, Haotong and Zhou, Chenxu and Wang, Weijie and Sun, Haiyang and Zhan, Kun and Lang, Xianpeng and Zhou, Xiaowei and Peng, Sida},
  booktitle={European Conference on Computer Vision},
  pages={156--173},
  year={2024},
  organization={Springer}
}

@article{chen2023periodic,
  title={Periodic vibration gaussian: Dynamic urban scene reconstruction and real-time rendering},
  author={Chen, Yurui and Gu, Chun and Jiang, Junzhe and Zhu, Xiatian and Zhang, Li},
  journal={arXiv preprint arXiv:2311.18561},
  year={2023}
}

@article{yang2023emernerf,
  title={Emernerf: Emergent spatial-temporal scene decomposition via self-supervision},
  author={Yang, Jiawei and Ivanovic, Boris and Litany, Or and Weng, Xinshuo and Kim, Seung Wook and Li, Boyi and Che, Tong and Xu, Danfei and Fidler, Sanja and Pavone, Marco and others},
  journal={arXiv preprint arXiv:2311.02077},
  year={2023}
}

@article{chen2024omnire,
  title={Omnire: Omni urban scene reconstruction},
  author={Chen, Ziyu and Yang, Jiawei and Huang, Jiahui and de Lutio, Riccardo and Esturo, Janick Martinez and Ivanovic, Boris and Litany, Or and Gojcic, Zan and Fidler, Sanja and Pavone, Marco and others},
  journal={arXiv preprint arXiv:2408.16760},
  year={2024}
}

@article{wang2024freevs,
  title={Freevs: Generative view synthesis on free driving trajectory},
  author={Wang, Qitai and Fan, Lue and Wang, Yuqi and Chen, Yuntao and Zhang, Zhaoxiang},
  journal={arXiv preprint arXiv:2410.18079},
  year={2024}
}

@article{guo2025dist,
  title={Dist-4d: Disentangled spatiotemporal diffusion with metric depth for 4d driving scene generation},
  author={Guo, Jiazhe and Ding, Yikang and Chen, Xiwu and Chen, Shuo and Li, Bohan and Zou, Yingshuang and Lyu, Xiaoyang and Tan, Feiyang and Qi, Xiaojuan and Li, Zhiheng and others},
  journal={arXiv preprint arXiv:2503.15208},
  year={2025}
}

@article{kerbl20233d,
  title={3D Gaussian splatting for real-time radiance field rendering.},
  author={Kerbl, Bernhard and Kopanas, Georgios and Leimk{\"u}hler, Thomas and Drettakis, George},
  journal={ACM Trans. Graph.},
  volume={42},
  number={4},
  pages={139--1},
  year={2023}
}

@inproceedings{wei2025omni,
  title={Omni-scene: Omni-gaussian representation for ego-centric sparse-view scene reconstruction},
  author={Wei, Dongxu and Li, Zhiqi and Liu, Peidong},
  booktitle={Proceedings of the Computer Vision and Pattern Recognition Conference},
  pages={22317--22327},
  year={2025}
}

@inproceedings{yang2023unisim,
  title={Unisim: A neural closed-loop sensor simulator},
  author={Yang, Ze and Chen, Yun and Wang, Jingkang and Manivasagam, Sivabalan and Ma, Wei-Chiu and Yang, Anqi Joyce and Urtasun, Raquel},
  booktitle={Proceedings of the IEEE/CVF Conference on Computer Vision and Pattern Recognition},
  pages={1389--1399},
  year={2023}
}

@inproceedings{zhao2025drivedreamer4d,
  title={Drivedreamer4d: World models are effective data machines for 4d driving scene representation},
  author={Zhao, Guosheng and Ni, Chaojun and Wang, Xiaofeng and Zhu, Zheng and Zhang, Xueyang and Wang, Yida and Huang, Guan and Chen, Xinze and Wang, Boyuan and Zhang, Youyi and others},
  booktitle={Proceedings of the Computer Vision and Pattern Recognition Conference},
  pages={12015--12026},
  year={2025}
}

@inproceedings{ni2025recondreamer,
  title={Recondreamer: Crafting world models for driving scene reconstruction via online restoration},
  author={Ni, Chaojun and Zhao, Guosheng and Wang, Xiaofeng and Zhu, Zheng and Qin, Wenkang and Huang, Guan and Liu, Chen and Chen, Yuyin and Wang, Yida and Zhang, Xueyang and others},
  booktitle={Proceedings of the Computer Vision and Pattern Recognition Conference},
  pages={1559--1569},
  year={2025}
}

@inproceedings{fan2025freesim,
  title={Freesim: Toward free-viewpoint camera simulation in driving scenes},
  author={Fan, Lue and Zhang, Hao and Wang, Qitai and Li, Hongsheng and Zhang, Zhaoxiang},
  booktitle={Proceedings of the Computer Vision and Pattern Recognition Conference},
  pages={12004--12014},
  year={2025}
}

@article{yang2024driving,
  title={Driving scene synthesis on free-form trajectories with generative prior},
  author={Yang, Zeyu and Pan, Zijie and Yang, Yuankun and Zhu, Xiatian and Zhang, Li},
  journal={arXiv preprint arXiv:2412.01717},
  year={2024}
}

@inproceedings{tian2025drivingforward,
  title={Drivingforward: Feed-forward 3d gaussian splatting for driving scene reconstruction from flexible surround-view input},
  author={Tian, Qijian and Tan, Xin and Xie, Yuan and Ma, Lizhuang},
  booktitle={Proceedings of the AAAI Conference on Artificial Intelligence},
  volume={39},
  number={7},
  pages={7374--7382},
  year={2025}
}

@inproceedings{wang2024drivedreamer,
  title={Drivedreamer: Towards real-world-drive world models for autonomous driving},
  author={Wang, Xiaofeng and Zhu, Zheng and Huang, Guan and Chen, Xinze and Zhu, Jiagang and Lu, Jiwen},
  booktitle={European conference on computer vision},
  pages={55--72},
  year={2024},
  organization={Springer}
}

@inproceedings{sun2020scalability,
  title={Scalability in perception for autonomous driving: Waymo open dataset},
  author={Sun, Pei and Kretzschmar, Henrik and Dotiwalla, Xerxes and Chouard, Aurelien and Patnaik, Vijaysai and Tsui, Paul and Guo, James and Zhou, Yin and Chai, Yuning and Caine, Benjamin and others},
  booktitle={Proceedings of the IEEE/CVF conference on computer vision and pattern recognition},
  pages={2446--2454},
  year={2020}
}

@inproceedings{geiger2012we,
  title={Are we ready for autonomous driving? the kitti vision benchmark suite},
  author={Geiger, Andreas and Lenz, Philip and Urtasun, Raquel},
  booktitle={2012 IEEE conference on computer vision and pattern recognition},
  pages={3354--3361},
  year={2012},
  organization={IEEE}
}

@article{gao2024vista,
  title={Vista: A generalizable driving world model with high fidelity and versatile controllability},
  author={Gao, Shenyuan and Yang, Jiazhi and Chen, Li and Chitta, Kashyap and Qiu, Yihang and Geiger, Andreas and Zhang, Jun and Li, Hongyang},
  journal={Advances in Neural Information Processing Systems},
  volume={37},
  pages={91560--91596},
  year={2024}
}

@article{blattmann2023stable,
  title={Stable video diffusion: Scaling latent video diffusion models to large datasets},
  author={Blattmann, Andreas and Dockhorn, Tim and Kulal, Sumith and Mendelevitch, Daniel and Kilian, Maciej and Lorenz, Dominik and Levi, Yam and English, Zion and Voleti, Vikram and Letts, Adam and others},
  journal={arXiv preprint arXiv:2311.15127},
  year={2023}
}

@inproceedings{barron2022mip,
  title={Mip-nerf 360: Unbounded anti-aliased neural radiance fields},
  author={Barron, Jonathan T and Mildenhall, Ben and Verbin, Dor and Srinivasan, Pratul P and Hedman, Peter},
  booktitle={Proceedings of the IEEE/CVF conference on computer vision and pattern recognition},
  pages={5470--5479},
  year={2022}
}

@inproceedings{barron2023zip,
  title={Zip-nerf: Anti-aliased grid-based neural radiance fields},
  author={Barron, Jonathan T and Mildenhall, Ben and Verbin, Dor and Srinivasan, Pratul P and Hedman, Peter},
  booktitle={Proceedings of the IEEE/CVF International Conference on Computer Vision},
  pages={19697--19705},
  year={2023}
}

@article{mildenhall2021nerf,
  title={Nerf: Representing scenes as neural radiance fields for view synthesis},
  author={Mildenhall, Ben and Srinivasan, Pratul P and Tancik, Matthew and Barron, Jonathan T and Ramamoorthi, Ravi and Ng, Ren},
  journal={Communications of the ACM},
  volume={65},
  number={1},
  pages={99--106},
  year={2021},
  publisher={ACM New York, NY, USA}
}

@article{muller2022instant,
  title={Instant neural graphics primitives with a multiresolution hash encoding},
  author={M{\"u}ller, Thomas and Evans, Alex and Schied, Christoph and Keller, Alexander},
  journal={ACM transactions on graphics (TOG)},
  volume={41},
  number={4},
  pages={1--15},
  year={2022},
  publisher={ACM New York, NY, USA}
}

@inproceedings{charatan2024pixelsplat,
  title={pixelsplat: 3d gaussian splats from image pairs for scalable generalizable 3d reconstruction},
  author={Charatan, David and Li, Sizhe Lester and Tagliasacchi, Andrea and Sitzmann, Vincent},
  booktitle={Proceedings of the IEEE/CVF conference on computer vision and pattern recognition},
  pages={19457--19467},
  year={2024}
}

@inproceedings{chen2024mvsplat,
  title={Mvsplat: Efficient 3d gaussian splatting from sparse multi-view images},
  author={Chen, Yuedong and Xu, Haofei and Zheng, Chuanxia and Zhuang, Bohan and Pollefeys, Marc and Geiger, Andreas and Cham, Tat-Jen and Cai, Jianfei},
  booktitle={European Conference on Computer Vision},
  pages={370--386},
  year={2024},
  organization={Springer}
}

@article{ho2020denoising,
  title={Denoising diffusion probabilistic models},
  author={Ho, Jonathan and Jain, Ajay and Abbeel, Pieter},
  journal={Advances in neural information processing systems},
  volume={33},
  pages={6840--6851},
  year={2020}
}

@article{song2020denoising,
  title={Denoising diffusion implicit models},
  author={Song, Jiaming and Meng, Chenlin and Ermon, Stefano},
  journal={arXiv preprint arXiv:2010.02502},
  year={2020}
}

@article{dhariwal2021diffusion,
  title={Diffusion models beat gans on image synthesis},
  author={Dhariwal, Prafulla and Nichol, Alexander},
  journal={Advances in neural information processing systems},
  volume={34},
  pages={8780--8794},
  year={2021}
}

@article{ho2022classifier,
  title={Classifier-free diffusion guidance},
  author={Ho, Jonathan and Salimans, Tim},
  journal={arXiv preprint arXiv:2207.12598},
  year={2022}
}

@inproceedings{rombach2022high,
  title={High-resolution image synthesis with latent diffusion models},
  author={Rombach, Robin and Blattmann, Andreas and Lorenz, Dominik and Esser, Patrick and Ommer, Bj{\"o}rn},
  booktitle={Proceedings of the IEEE/CVF conference on computer vision and pattern recognition},
  pages={10684--10695},
  year={2022}
}

@article{ho2022video,
  title={Video diffusion models},
  author={Ho, Jonathan and Salimans, Tim and Gritsenko, Alexey and Chan, William and Norouzi, Mohammad and Fleet, David J},
  journal={Advances in neural information processing systems},
  volume={35},
  pages={8633--8646},
  year={2022}
}

@article{hu2024metric3d,
  title={Metric3d v2: A versatile monocular geometric foundation model for zero-shot metric depth and surface normal estimation},
  author={Hu, Mu and Yin, Wei and Zhang, Chi and Cai, Zhipeng and Long, Xiaoxiao and Chen, Hao and Wang, Kaixuan and Yu, Gang and Shen, Chunhua and Shen, Shaojie},
  journal={IEEE Transactions on Pattern Analysis and Machine Intelligence},
  year={2024},
  publisher={IEEE}
}

@article{li2024bevformer,
  title={Bevformer: learning bird's-eye-view representation from lidar-camera via spatiotemporal transformers},
  author={Li, Zhiqi and Wang, Wenhai and Li, Hongyang and Xie, Enze and Sima, Chonghao and Lu, Tong and Yu, Qiao and Dai, Jifeng},
  journal={IEEE Transactions on Pattern Analysis and Machine Intelligence},
  year={2024},
  publisher={IEEE}
}

@inproceedings{huang2023tri,
  title={Tri-perspective view for vision-based 3d semantic occupancy prediction},
  author={Huang, Yuanhui and Zheng, Wenzhao and Zhang, Yunpeng and Zhou, Jie and Lu, Jiwen},
  booktitle={Proceedings of the IEEE/CVF conference on computer vision and pattern recognition},
  pages={9223--9232},
  year={2023}
}

@inproceedings{zhang2018unreasonable,
  title={The unreasonable effectiveness of deep features as a perceptual metric},
  author={Zhang, Richard and Isola, Phillip and Efros, Alexei A and Shechtman, Eli and Wang, Oliver},
  booktitle={Proceedings of the IEEE conference on computer vision and pattern recognition},
  pages={586--595},
  year={2018}
}

@inproceedings{ren2025gen3c,
  title={Gen3c: 3d-informed world-consistent video generation with precise camera control},
  author={Ren, Xuanchi and Shen, Tianchang and Huang, Jiahui and Ling, Huan and Lu, Yifan and Nimier-David, Merlin and M{\"u}ller, Thomas and Keller, Alexander and Fidler, Sanja and Gao, Jun},
  booktitle={Proceedings of the Computer Vision and Pattern Recognition Conference},
  pages={6121--6132},
  year={2025}
}

@misc{Seitzer2020FID,
  author={Maximilian Seitzer},
  title={{pytorch-fid: FID Score for PyTorch}},
  month={August},
  year={2020},
  note={Version 0.3.0},
  howpublished={\url{https://github.com/mseitzer/pytorch-fid}},
}

@misc{stylegan_v,
    title={StyleGAN-V: A Continuous Video Generator with the Price, Image Quality and Perks of StyleGAN2},
    author={Ivan Skorokhodov and Sergey Tulyakov and Mohamed Elhoseiny},
    journal={arXiv preprint arXiv:2112.14683},
    year={2021}
}

@article{wang2004image,
  title={Image quality assessment: from error visibility to structural similarity},
  author={Wang, Zhou and Bovik, Alan C and Sheikh, Hamid R and Simoncelli, Eero P},
  journal={IEEE transactions on image processing},
  volume={13},
  number={4},
  pages={600--612},
  year={2004},
  publisher={IEEE}
}
}


\end{document}